\documentclass[11pt,a4paper]{article}
\usepackage[hyperref]{eacl2021}
\usepackage{times}
\usepackage{latexsym}

\usepackage{xspace}
\usepackage{xargs}
\usepackage{xcolor}
\usepackage[colorinlistoftodos,prependcaption,textsize=tiny]{todonotes}
\usepackage{stfloats}

\usepackage{marginnote}
\usepackage{booktabs}
\usepackage{amsmath}
\usepackage{amssymb}
\usepackage{hyperref}
\usepackage{appendix}
\usepackage{multirow}
\usepackage{arydshln}
\usepackage{multirow}
\usepackage{tabularx,booktabs}
\usepackage{url}
\usepackage{framed}
\definecolor{shadecolor}{RGB}{150,150,150}
\definecolor{mygreen}{rgb}{0.13, 0.55, 0.13}

\usepackage{microtype}
\usepackage[final]{changes}

\aclfinalcopy % Uncomment this line for the final submission
\newcommand\baseline{{RoBERTa}}

\newcommand\baselineplus{{RoBERTa++}}
\newcommand\ourmodel{{TEK$_{PF}$}}
\newcommand\ourmethod{{TEK}}
\newcommand\ourpretraining{{TEK}}

\newcommand\baselinepluspretraining{{Context-O}}
\newcommand\ourfinetuning{{TEK}}
\newcommand\baselinefinetuning{{Context-O}}
\newcommand{\ignore}[1]{}

\newcommand\tf[1]{\textbf{#1}}

\newcommand\cls{\texttt{[CLS]}}
\newcommand\sep{\texttt{[SEP]}}
\newcommand{\eat}[1]{\ignorespaces}

\newcounter{iu}

\title{Contextualized Representations Using Textual Encyclopedic Knowledge}
\author{Mandar Joshi\thanks{~~Work completed while interning at Google} $~^{\dagger}$ \quad Kenton Lee $^{\epsilon}$ \quad Yi Luan $^{\epsilon}$ \quad Kristina Toutanova $^{\epsilon}$ \\[4pt]
$^{\dagger}$ Allen School of Computer Science \& Engineering, University of Washington, Seattle, WA \\
{\tt \{mandar90\}@cs.washington.edu}\\[4pt]
$^{\epsilon}$ Google Research, Seattle\\
{\tt \{kentonl,luanyi,kristout\}@google.com} \\[4pt]
}
\date{}

\begin{document}
\maketitle
 \begin{abstract} 
We present a method to represent input texts by contextualizing them jointly with dynamically retrieved textual encyclopedic background knowledge from multiple documents.
 We apply our method to reading comprehension tasks by  encoding questions and passages together with background sentences about the entities they mention.
 We show that integrating background knowledge directly from text is  effective for tasks focusing on factual reasoning and allows direct reuse of powerful pretrained BERT-style encoders. Moreover, knowledge integration can be further improved with suitable pretraining via a self-supervised masked language model objective over words in background-augmented input text. On TriviaQA, our approach obtains improvements of 1.6 to 3.1 F1 over comparable RoBERTa models which do not integrate background knowledge dynamically. On MRQA, a large collection of diverse question answering datasets, we see consistent gains in-domain along with  large improvements out-of-domain on BioASQ (2.1 to 4.2 F1), TextbookQA (1.6 to 2.0 F1), and DuoRC (1.1 to 2.0 F1). 
%  We will release code and models.
%  \footnote{Our code is available at \href{AnonymizedLink}{AnonymizedLink}} 
 \end{abstract}
 
\section{Introduction}

%We present methods to represent input texts by contextualizing them jointly with dynamically retrieved textual background knowledge, in the form of sentences from multiple documents in Wikipedia. We apply the methods on a staple language understanding task, reading comprehension, which often requires such background knowledge (Figure~\ref{fig:intro_figure}).

% \begin{figure}[t]
% \small
% \begin{tabular}{p{7cm}}
% \toprule
% \textbf{Question}: What is the \emph{pen-name of the gossip columnist in the Daily Express}, first written by Tom Driberg in 1928 and later Nigel Dempster in the 1960s? \\
% \textbf{Answer}: William Hickey\\
% \textbf{Context}: Nigel Richard Patton Dempster was a British journalist...Best known for his celebrity gossip columns, his work appeared in the Daily Express and Daily Mail... he was a contributor to the  \textbf{William Hickey} column... \\

% \textbf{Background}: \textcolor{blue}{
% ``William Hickey'' is the \emph{pseudonymous byline of a gossip column published in the Daily Express}.}
% \\
% \bottomrule
% \end{tabular}
%     \caption{An example from TriviaQA showing how background sentences from Wikipedia help define the meaning of phrases in the context and their relationship to phrases in the question. The answer \emph{William Hickey} is connected to the question phrase \emph{pen-name of the gossip columnist in the Daily Express} through the background sentence.}
%     \label{fig:intro_figure}
% \end{figure}

\begin{figure}[!t]
\centering
\includegraphics[scale=0.30]{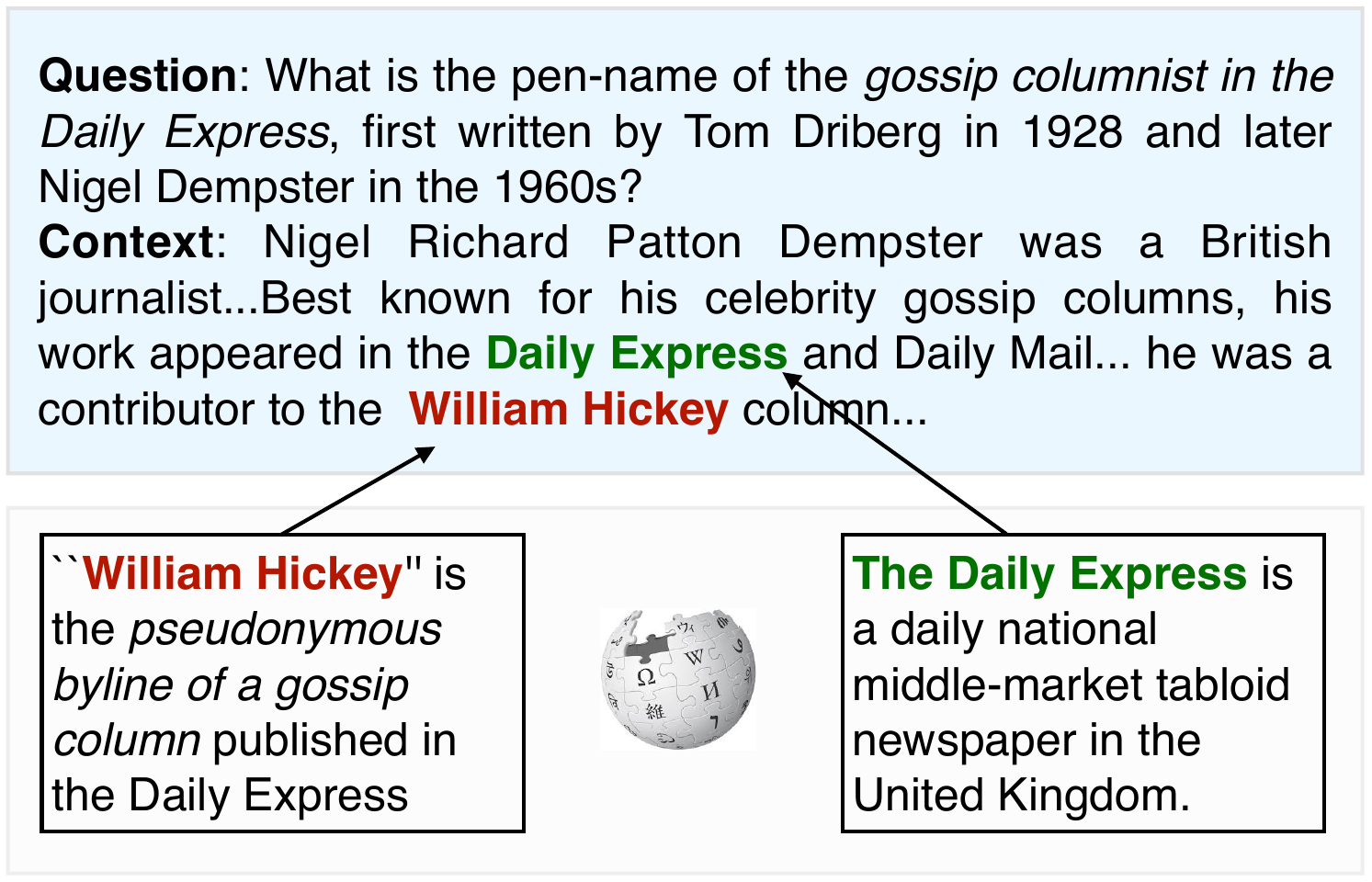}
\caption{A TriviaQA example showing how background sentences from Wikipedia help define the meaning of phrases in the context and their relationship to phrases in the question. The answer \emph{William Hickey} is connected to the question phrase \emph{pen-name of the gossip columnist in the Daily Express} through the background sentence.
}
\label{fig:intro_figure}
\end{figure}

\begin{figure*}[!t]
\centering
\includegraphics[width=0.9\textwidth, keepaspectratio, trim={5.2cm 9.1cm 7.1cm 12.3cm}, clip]{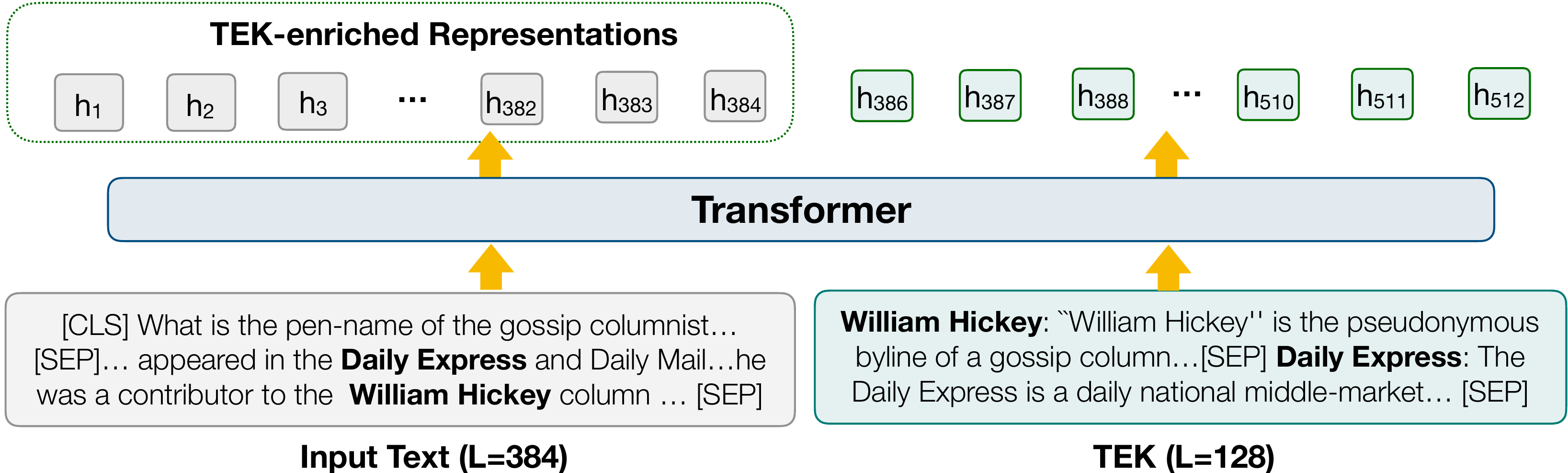}
\caption{We contextualize the input text, in this case a question and a passage, together with textual encyclopedic knowledge (TEK) using a pretrained Transformer to create TEK-enriched representations.}
\label{fig:encoding}
\end{figure*}

% \begin{figure*}[!t]
% \centering
% \includegraphics[scale=0.30]{encoding.pdf}
% \caption{We contextualize the input text, in this case a question and a passage, together with textual encyclopedic knowledge (TEK) using a pretrained Transformer to create TEK-enriched representations.}
% \label{fig:encoding}
% \end{figure*}

% Why is the problem hard and what have other people done
Current self-supervised representations, trained at large scale from document-level contexts, are known to encode linguistic~\cite{tenney2019you} and factual~\cite{petroni2019language} knowledge into their
% deep neural network 
parameters. 
Yet, even large pretrained representations  are unable to capture and preserve all factual knowledge they have ``read'' during pretraining  due to the long tail of entity and event-specific information \cite{logan-etal-2019-baracks}. 
% \added{
For open-domain tasks, where the input consists of only a question or a statement out of context, as in open-domain QA  or factuality prediction, previous work has retrieved and used text that may contain or entail the needed answer to build representations for the task ~\cite{drqa, guu2020realm, lewis2020retrievalaugmented,oh2017multi,kadowaki-etal-2019-event}. 

On the other hand, when relevant text is provided as input, such as in reading comprehension tasks~\cite{rajpurkar2016squad}, relation extraction, syntactic analysis, etc.,  which can be cast as tasks of labeling spans in the input text, prior work has not focused on drawing background information from external text sources. Instead, most research has explored architectures to integrate background  from structured knowledge bases to form input text representations ~\cite{bauer-etal-2018-commonsense,mihaylov-frank-2018-knowledgeable,yang-etal-2019-enhancing-pre,zhang2019ernie,peters-etal-2019-knowledge}.\footnote{ A notable exception is \newcite{weissenborn2017dynamic}, with a specialized architecture which uses \emph{textual} entity descriptions. }
% }

%~\cite{weissenborn2017dynamic,bauer-etal-2018-commonsense,mihaylov-frank-2018-knowledgeable,yang-etal-2019-enhancing-pre,zhang2019ernie,peters-etal-2019-knowledge}. 
% While integrating structured knowledge can enable precise inference, the coverage of knowledge in structured resources is often limited.
%~\cite{bauer-etal-2018-commonsense,mihaylov-frank-2018-knowledgeable,yang-etal-2019-enhancing-pre,zhang2019ernie,peters-etal-2019-knowledge}. While integrating structured knowledge can enable precise inference, the coverage of knowledge in structured resources is often limited.

%%~\footnote{Except \newcite{weissenborn2017dynamic} who used Wikipedia abstracts with a specialized architecture.}

We posit that representations should be able to directly integrate \emph{textual} background knowledge since a wider scope of information is more readily available in textual form. Our method represents input texts by jointly encoding them with \emph{dynamically retrieved} sentences from the Wikipedia pages of entities they mention. We term these representations TEK-enriched, for Textual Encyclopedic Knowledge  (Figure~\ref{fig:encoding} shows an illustration), and use them for reading comprehension (RC) by contextualizing questions and passages together with  retrieved Wikipedia background sentences. 
% \added{Text retrieval is commonly used for problems such as open-domain question answering~\cite{drqa} and event causality recognition~\cite{oh2017multi,kadowaki-etal-2019-event}, where contexts addressing the required information need are not given as input; we show that integrating textual background knowledge is useful even when high quality evidence passages are available in reading comprehension.}
Such background knowledge can help reason about the relationships between questions and passages. Figure~\ref{fig:intro_figure} shows an example question from the TriviaQA dataset ~\cite{joshi-etal-2017-triviaqa} asking for \emph{the pen-name of a gossip columnist}. Encoding relevant background knowledge (\emph{pseudonymous byline of a gossip column published in the Daily Express}) helps ground the vague reference to \emph{the William Hickey column} in the given document context.

% Using dynamically retrieved text as background knowledge allows us to directly reuse powerful pretrained BERT-style encoders~\cite{devlin-etal-2019-bert}. We show that an off-the-shelf RoBERTa~\cite{liu2019roberta} model can be directly applied to minimally structured TEK-enriched input encoding schemes, which allow the encoder to distinguish between the original context and background sentences. The model can then be successfully finetuned to considerably improve on current state-of-the art methods which only consider context from a single document (Section~\ref{sec:results}). The improvement comes without an increase in the length of the input  window for the Transformer~\cite{vaswani2017attention}.
% % (see Section~\ref{sec:context_vs_background}).
%dynamically retrieved
Using  text as background knowledge allows us to directly reuse powerful pretrained BERT-style encoders~\cite{devlin-etal-2019-bert}. We show that an off-the-shelf RoBERTa~\cite{liu2019roberta} model can be directly finetuned on minimally structured TEK-enriched inputs, which are formatted to allow the encoder to distinguish between the original passages and background sentences. This method considerably improves on current state-of-the art methods which only consider context from a single \added{input} document (Section~\ref{sec:results}). The improvement comes without an increase in the length of the input  window for the Transformer~\cite{vaswani2017attention}.
% (see Section~\ref{sec:context_vs_background}).

% Although existing state-of-the-art models pretrained on contiguous segments of  document text (e.g. RoBERTa), can be used to provide a good starting point for task-specific TEK-enriched representations, there is a mismatch between the type of input seen during pretraining (single document segments) and  the type of input the model is asked to represent for downstream tasks (document text with background Wikipedia sentences from multiple pages). We therefore design a suitable self-supervised large-scale pretraining objective, which drives learning of token representations in TEK-enriched input, and show that this better matched pretraining task brings further substantial gains.

Although existing pretrained models provide a good starting point for task-specific TEK-enriched representations, there is still a mismatch between the type of input seen during pretraining (single document segments) and  the type of input the model is asked to represent for downstream tasks (document text with background Wikipedia sentences from multiple pages). 
% We therefore design a suitable large-scale self-supervised  pretraining task, for learning token representations in the TEK-enriched input, and show that this better matched pretraining task brings further substantial gains.
We show that the Transformer model can be substantially improved by reducing this mismatch via self-supervised  masked language model (MLM)~\cite{devlin-etal-2019-bert} pretraining on TEK-augmented input texts.

Our approach records considerable improvements over state of the art base (12-layer) and large (24-layer) Transformer models for in-domain and out-of-domain document-level extractive question answering (QA), for tasks where factual knowledge about entities is important and well-covered by the background collection. On TriviaQA, we see improvements of 1.6 to 3.1 F1, respectively, over comparable RoBERTa models which do not integrate background information. On MRQA~\cite{fisch2019mrqa}, a large collection of diverse QA datasets, we see consistent gains in-domain along with large improvements out-of-domain on BioASQ (2.1 to 4.2 F1), TextbookQA (1.6 to 2.0 F1), and DuoRC (1.1 to 2.0 F1). 
% We will release code and models.

\section{\ourmethod-enriched Representations}
%\subsection{Background}
% BERT \cite{devlin-etal-2019-bert} is a self-supervised approach that pre-trains a deep transformer encoder \cite{vaswani2017attention}, which is then finetuned for a particular downstream task. The input to BERT consists of two sequences ($X_A , X_B$). The sequence $X_A$ is some sequence of tokens from the corpus; $X_B$ is (1) either the following sequence from the same document, or (b) a sequence randomly sampled from a different document in the corpus. The two sequences are separated by a special \sep~token, and the entire input is preprocessed by replacing 15\% of randomly selected tokens with placeholder tokens. BERT optimizes two training objectives -- masked language model (MLM) and next sentence prediction (NSP). The MLM task predicts masked tokens in the input from the output representations of their placeholder tokens. The NSP task predicts whether $X_B$ is a continuation of $X_A$ from the same document.
% \luanyi{Too much details on BERT maybe? Especially too much on NSP since we are not using it. Do we even need to make this a separate 2.1 Background section? Maybe just an opening paragraph would be enough to introduce our approach?}

We follow recent work on pretraining bidirectional Transformer representations  on unlabeled text, and finetuning them for downstream tasks~\cite{devlin-etal-2019-bert}. Subsequent approaches have shown significant improvements over BERT by improving the training example generation, the masking strategy,  the pretraining objectives, and the optimization methods~\cite{liu2019roberta,joshi2019spanbert}. We build on these improvements to train TEK-enriched representations and use them for extractive QA.
%In the next section, we will present our approach to improve the model's ability to integrate background information while building upon recent improvements like span based masking~\cite{joshi2019spanbert} and large scale pretraining~\cite{liu2019roberta}. 

% \subsection{\ourmethod~Enriched Representations}
% \luanyi{Overall suggestion for this section: We probably need to re-orgnize the structure which is now a little redundant now. How about: 3 subsections (or paragraphs) Sec 2.2: 1) Background Knowledge Retrieval 2)Modeling 3)Discussion. Explain difference between pretraining and finetuning in each subsections.}
Our approach seeks to contextualize input text $X = ( x_1 , \ldots , x_n )$ jointly with relevant textual encyclopedic background knowledge $B$ retrieved dynamically from multiple documents.  We define a retrieval function, $f_{ret}(X, \mathcal{D})$, which takes $X$ as input and retrieves a list of text spans $B = ( B_1 , \ldots , B_M )$ from the corpus $\mathcal{D}$. In our implementation, each of the text spans $B_i$ is a sentence. 
    The encoder then represents $X$ by jointly encoding $X$ with $B$ using $f_{enc}(X, B)$ such that the output representations of $X$ are cognizant of the information present in $B$ (see Figure~\ref{fig:encoding}). We use a deep Transformer encoder operating over the input sequence \cls $X$ \sep $B$ \sep for  $f_{enc}(\cdot)$.
%during pretraining which is then finetuned on RC data.
%The Transformer takes as input the sequence \cls $X$ \sep $B$ \sep. 

We refer to inputs $X$ generically as \emph{contexts}. These could be either contiguous word sequences from documents (passages), or, for the QA application, question-passage pairs, which we refer to as \emph{RC-contexts}. For a fixed Transformer input length limit (which is necessary for computational efficiency), there is a trade-off between the length of the document context (the length of $X$) and the amount of background knowledge (the length of $B$). Section~\ref{sec:context_vs_background} explores this trade-off and shows that for an encoder input limit of 512, the values of $N_C = 384$ for the length of $X$ and $N_B = 128$ for the length of $B$ provide an effective compromise.

We use a simple implementation of the background retrieval function $f_{ret}(X, \mathcal{D})$, using an entity linker for finetuning (Section~\ref{sec:finetuning-method}) and Wikipedia hyperlinks for pretraining (Section~\ref{sec:pretraining-method}), and a way to score the relevance of individual sentences using \emph{ngram} overlap. 
% We begin by specifying the retrieval function we use for RC-contexts followed by the QA model which uses TEK-enriched representations in Section~\ref{sec:finetuning-method}. 
% We then detail baseline and newly proposed self-supervised pretraining methods for our representations in Section~\ref{sec:pretraining-method}.

\subsection{\ourmethod-Enriched Question Answering}
\label{sec:finetuning-method}

The input $X$ for the extractive QA task consists of the question $Q$ and a candidate passage $P$. We use the following retrieval function $f_{ret}(X)$ to obtain relevant background $B$.
\paragraph{Background Knowledge Retrieval for QA}
We detect entity mentions in $X$ using a proprietary Wikipedia-based entity linker,\footnote{We also report results on publicly available linkers showing that our method is robust to the exact choice of the linker (Section~\ref{sec:ablation}).} and form a candidate pool of background segments $B_i$ as the union of the sentences in the Wikipedia pages of the detected entities. These sentences are then ranked based on their number of overlapping ngrams with the question (equally weighted unigrams, bigrams, and trigrams).
% \begin{enumerate}
%     \item We detect entity mentions in $X$ using a proprietary Wikipedia-based entity linker.
%     % {~\footnote{\added{Work in progress aims to evaluate the impact of the entity linker and compare to publicly available resources.}}}
%     % \footnote{We use the Google Cloud Natural Language API, extracting the ``entity analysis'' results -- \url{https://cloud.google.com/natural-language/docs/basics\#entity analysis}}.
%     \item We form a candidate pool of background segments $B_i$ as the union of the sentences in the Wikipedia pages of the detected entities. We rank the sentences based on their number of overlapping ngrams with the question (equally weighted unigrams, bigrams, and trigrams).
% \end{enumerate}
To form the input for the Transformer encoder, each background sentence is minimally structured as $B_i$ by prepending the name of the entity whose page it belongs to along with a separator `:' token. Each sentence $B_i$ is followed by \sep.
% ~(Appendix~\ref{sec:input_examples} for examples). 
Appendix~\ref{sec:input_examples} shows an example of an RC-context with background knowledge segments.
% \luanyi{Bring reference to the table earlier in the section to make it more visible to readers?}

\paragraph{QA Model} Following BERT, our QA model architecture consists of two independent linear classifiers for predicting the answer span boundary (start and end) on top of the output representations of $X$. We assume that the answer, if present, is contained only in the given passage, $P$, and do not consider potential mentions of the answer in the background $B$.
For instances which do not contain the answer, we set the answer span to be the special token \cls. We use a fixed Transformer input window size of $512$, and use a sliding window with a stride of $128$ tokens to handle longer  documents.  Our TEK-enriched representations use document passages of length $384$ while baselines use longer passages of length $512$.
% moved to details
%For TriviaQA only, we do not use the full input documents, but a concatenation of the first four 400-token passages selected by the linear passage ranker of \citet{clark-gardner-2018-simple}.

\subsection{TEK-enriched Pretraining}
\label{sec:pretraining-method}
% We integrate textual background information about entities by pretraining representations which encode contexts together with their background Wikipedia sentences.
Standard pretraining uses contiguous document-level natural language inputs. Since TEK-augmented inputs are formatted as natural language sequences, off-the-shelf pretrained models can be used as a starting point for creating TEK-enriched representations. As one of our approaches, we use a standard single-document pretraining model. 
% \kristout{I don't think we use RoBERTa because we use RoBERTa++}\mj{Changed}

While the input format is the same, there is a mismatch between contiguous document segments and TEK-augmented inputs sourced from multiple documents. We propose an additional pretraining stage---starting from the RoBERTa parameters, we resume pretraining using an MLM  objective on TEK-augmented document text $X$, which encourages the model to integrate the knowledge from multiple background segments. 
% The new task is defined on TEK-augmented document text $X$, and is detailed below.

\paragraph{Background Knowledge Retrieval in Pretraining}
% \luanyi{Added a paragraph title to make it parallel to \textbf{Model}} 
In pretraining, $X$ is a contiguous block of text from Wikipedia. The retrieval function $f_{ret}(X, \mathcal{D})$ returns $B = ( B_1 , \ldots , B_M )$ where each $B_i$ is a sentence from the Wikipedia page of some entity hyperlinked from a span in $X$. 
% \luanyi{notation description is redundant from the intro paragraph of this section.}
We use high-precision Wikipedia hyperlinks instead of an entity linker for pretraining.
% \luanyi{by pre-pending the title of its Wikipedia page. Is it Wikipedia title or the entity surface form?}\mj{Title}  
The background candidate sentences are ranked by their  ngram overlap with  $X$. The top ranking sentences in $B$ up to $N_B$ tokens are used. If no entities are found in $X$,  $B$ is constructed from the context following $X$ from the same document.
%without any additional preprocessing. 

\paragraph{Training Objective} We continue pretraining a deep Transformer using the MLM objective~\cite{devlin-etal-2019-bert}  after initializing the parameters with pretrained RoBERTa weights. Following improvements in SpanBERT~\cite{joshi2019spanbert}, we mask spans with lengths sampled from a geometric distribution in the entire input ($X$ and $B$). We use a single segment ID, and remove the next sentence prediction objective which has been shown to not improve performance~\cite{joshi2019spanbert,liu2019roberta} for multiple tasks including QA. 
% To evaluate the impact of the new TEK-augmented pretraining method while controlling for the number of steps and other pretraining hyperparameters, we also experiment with RoBERTa extended with the same number of additional pretraining steps and masking strategy (termed RoBERTa++).
We evaluate two methods building textual-knowledge enriched representations for QA differing in the pretraining approach used:
\paragraph{TEK$_{PF}$} Our full approach TEK$_{PF}$~\footnote{The subscripts $P$ and $F$ stand for pretraining and finetuning, respectively. } consists of two stages: (a) 200K steps of TEK-pretraining on Wikipedia starting from the RoBERTa checkpoint, and (b) finetuning and doing inference on RC-contexts augmented with TEK background.
\paragraph{TEK$_{F}$} TEK$_{F}$ replaces the first specialized pretraining stage in TEK$_{PF}$ with 200K steps for standard single-document-context pretraining for a fair comparison with \ourmodel, but follows the same finetuning regimen. 

\section{Experimental Setup}

\begin{table}
\centering
\small
\begin{tabular}{l @{\hspace{0.5cm}} r r r r}
\toprule
Task & Train & Dev & Test \\
\midrule
TQA Wiki & 61,888 & 7,993 & 7,701\\
TQA Web & 528,979 & 68,621 & 65,059\\
MRQA & 616,819 & 58,221 & 9,633\\
\bottomrule
\end{tabular}
\caption{Data statistics for TriviaQA and MRQA. }
\label{tab:task_setup}
\end{table}

We perform experiments on TriviaQA and MRQA, two large extractive question answering benchmarks (see Table~\ref{tab:task_setup} for dataset statistics).

\paragraph{TriviaQA} TriviaQA~\cite{joshi-etal-2017-triviaqa} contains trivia questions paired with evidence collected via entity linking and web search. 
The dataset is \emph{distantly supervised} in that the answers are contained in the evidence but the context may not support answering the questions. 
We experiment with both the Wikipedia and Web tasks.

% For this benchmark, we follow the input preprocessing of \citet{clark-gardner-2018-simple}. The input to our model is the concatenation of the first four 400-token passages selected by their linear passage ranker.  For training, we define
% the gold span to be the first occurrence of the gold
% answer(s) in the context~\cite{joshi-etal-2017-triviaqa,talmor-berant-2019-multiqa}. 

\paragraph{MRQA} The MRQA shared task~\cite{fisch2019mrqa} consists of several widely used QA datasets unified into a common format aimed at evaluating out-of-domain generalization. The data consists of a training set, in-domain and out-of-domain dev sets, and a private out-of-domain test set. The training and the in-domain dev sets  consist of modified versions of corresponding sets from SQuAD~\cite{rajpurkar2016squad}, NewsQA~\cite{trischler2017newsqa}, SearchQA~\cite{dunn2017searchqa}, TriviaQA Web~\cite{joshi-etal-2017-triviaqa}, HotpotQA~\cite{yang2018hotpotqa} and Natural Questions~\cite{kwiatkowski2019natural}. The out-of-domain test evaluation, including access to questions and passages, is only available through Codalab.
% \footnote{\url{https://worksheets.codalab.org/worksheets/0x926e37ac8b4941f793bf9b9758cc01be/}}
Due to the complexity of our system which involves entity linking and retrieval, we perform development and model selection on the in-domain dev set and treat the out-of-domain dev set as the test set. The out-of-domain set we evaluate on has examples from BioASQ~\cite{tsatsaronis2015overview}, DROP~\cite{dua-etal-2019-drop}, DuoRC~\cite{saha-etal-2018-duorc}, RACE~\cite{lai-etal-2017-race}, RelationExtraction~\cite{levy-etal-2017-zero}, and TextbookQA~\cite{Kembhavi2017tqa}.

% Table~\ref{tab:task_setup} shows the number of training, development, and test examples for each task, with TriviaQA Web (dev and test) and MRQA in-domain dev having well over 55K examples. 

% \subsection{Implementation}
% We implemented all models in TensorFlow~\cite{tensorflow2015-whitepaper}. For pretraining, we used the 12-layer RoBERTa-base and 24-layer RoBERTa-large configurations, and initialized the parameters from their respective checkpoints. \added{In TEK-augmented pretraining,} we further pretrained the models for 200K steps with a batch size of $512$ and BERT's triangular learning rate schedule with a warmup of $5000$ steps \added{on TEK-augmented contexts}. We used a peak learning rate of $0.0001$ for base and $5e^{-5}$ for large models. For finetuning hyperparameters, see Appendix~\ref{sec:finetune_procedure}. 

\subsection{Baselines}
% Our full approach TEK$_{PF}$ consists of two stages: (a) 200K steps of TEK-pretraining on Wikipedia starting from the RoBERTa checkpoint, and (b) finetuning and doing inference on RC-contexts augmented with TEK background. \added{TEK$_{F}$ replaces the first specialized pretraining stage with 200K steps for standard single-document-context pretraining for a fair comparison with \ourmodel,\kristout{This should be in the method section} but follows the same finetuning regimen.} 
We compare \ourmodel~and TEK$_{F}$ with two baselines, \baseline~and \baselineplus. Both use the same architecture as our approach, but use only original RC-contexts for finetuning and inference, and use standard single-document-context RoBERTa pretraining. \ourmodel~and TEK$_{F}$ use $N_C=384$ and $N_B=128$, while both baselines use $N_C=512$ and $N_B=0$. 
% \begin{itemize}
\paragraph{\baseline} We finetune the model on QA data without knowledge augmentation starting from the same RoBERTa checkpoint that is used as an initializer for TEK-augmented pretraining. 
% The suffix $PF$ indicates that both \tf{p}retraining and \tf{f}inetuning stages used only contexts (contiguous text sequences from documents).

\paragraph{\baselineplus} For a fair evaluation of the new TEK-augmented pretraining method while controlling for the number of pretraining steps and other hyperparameters, we extend RoBERTa's pretraining for an additional 200K steps on single contiguous blocks of text (without background information). We use the same masking and other hyperparameters as in TEK-augmented pretraining. This pretrained checkpoint is also used to initialize parameters for our TEK$_F$ approach.

The implementation details of all models, including hyperpameters, can be found in Appendix~\ref{sec:finetune_procedure}.

% \section{Experiments}

\section{Results}
\label{sec:results}

\begin{table}[!t]
  \centering
  \small
  \begin{tabular}{l @{\hspace{0.5cm}} c c c c c}
    \toprule
     & \multicolumn{2}{c}{TQA Wiki} & \multicolumn{2}{c}{TQA Web} \\
     & EM & F1 & EM & F1 \\
    \midrule
    \multicolumn{5}{c}{Previous work} \\
    \midrule
    \citet{clark-gardner-2018-simple} & 64.0 & 68.9 & 66.4 & 71.3\\
    \citet{weissenborn2017dynamic} & 64.6 & 69.9 & 67.5 &	72.8 \\
    \citet{wang-etal-2018-multi-granularity} & 66.6 & 71.4 & 68.6 & 73.1 \\
    \citet{LewisTQA} & 67.3 & 72.3 & - & - \\
    \midrule
    \multicolumn{5}{c}{This work} \\
    \midrule
    \baseline~(Base) & 66.7 &	71.7 &	77.0 &	81.4 \\
    \baselineplus~(Base) & 68.0 &	72.9 &	76.8 &	81.4\\
    TEK$_F$ (Base) & 70.0  & 74.8 & 78.2 &	83.0\\
    \ourmodel~(Base) & \tf{71.2} &	\tf{76.0} &	\tf{78.8} &	\tf{83.4}\\
    \midrule
    \baseline~(Large) & 72.3 &	76.9 &	80.6 &	85.1 \\
    \baselineplus~(Large) & 72.9 &	77.5 &	81.1 &	85.5 \\
    TEK$_F$(Large) & 74.1  & 78.6	 &	82.2 &	86.5\\
    \ourmodel~(Large) &  \tf{74.6} &	\tf{79.1} &	\tf{83.0} &	\tf{87.2}\\
    \bottomrule
  \end{tabular}
  \caption{Test set performance on TriviaQA.}
  \label{tab:main_results_tqa}
\end{table}

\begin{table*}[!t]
    \centering
    \small
    \begin{tabular}{l @{\hspace{0.5cm}} c c c c c c c c}
        \toprule
         & {MRQA-In} & BioASQ & TextbookQA & DuoRC & RE & DROP & RACE & {MRQA-Out}\\
        \midrule
        \multicolumn{8}{c}{Shared task} \\
        \midrule
        D-Net (Ensemble) & 84.82 & - & - & - & - & - & - & 70.42 \\
        Delphi & - & 71.98 & 65.54 & 63.36 & 87.85 & 58.9 & 53.87 & 66.92 \\
        \midrule
        \multicolumn{8}{c}{This work} \\
        \midrule
        \baseline~(Base) & 82.98 & 68.80 & 58.32 & 62.56 & 86.87 & \tf{54.88} & \tf{49.14} & 68.17 \\
        \baselineplus~(Base) & 83.22 & 68.36 & 60.51 & 62.40 & 87.93 & 53.11 & 47.90 & 68.38\\
        TEK$_F$ (Base) & 83.44 & 69.71 & 62.19 & 63.43 & 87.49 & 51.04 & 46.43 & 68.46\\
        \ourmodel~(Base) & \tf{83.71} & \tf{72.58} & \tf{62.55} & \tf{64.43} & \tf{88.29} & 54.58 & 47.75 & \tf{70.01}\\
        \midrule
        \baseline~(Large) & 85.75 & 73.41 & 65.95 & 66.79 & 88.82 & \tf{68.63} & \tf{56.84} & 74.02\\
        \baselineplus~(Large) & 85.80 & 74.73 & 67.51 & 67.40 & \tf{89.58} & 67.62 & 55.95 & 74.58 \\
        TEK$_F$ (Large) & 86.23 & 75.37 & 68.17 & \tf{68.80} & 89.43 & 67.46 & 55.20 & 74.88\\
        \ourmodel~(Large) & \tf{86.33} & \tf{76.80} & \tf{69.10} & {68.54} & 89.15 & 66.24 & 56.14 & \tf{75.00} \\
    \bottomrule
  \end{tabular}
  \caption{In-domain and out-of-domain performance (F1) on MRQA. RE refers to the Relation Extraction dataset. MRQA-Out refers to the averaged out-of-domain F1.}
  \label{tab:main_results_mrqa}
\end{table*}
% We evaluate our approach first on the Wikipedia and Web tasks of TriviaQA, followed by in-domain and out-of-domain evaluation on MRQA.   

% Table ~\ref{tab:main_results_tqa} shows that our approach, \ourmodel, outperforms both baselines by a considerable margin on TriviaQA. 
\paragraph{TriviaQA} Table ~\ref{tab:main_results_tqa} compares our approaches with baselines and previous work. The 12-layer variant of our \baseline~baseline outperforms or matches the performance of several previous systems including ELMo-based ones ~\cite{wang-etal-2018-multi-granularity,LewisTQA} which are specialized for this task. We also see that \baselineplus~outperforms \baseline, indicating that there is still room for improvement by simply pretraining for more steps on task-domain relevant text. Furthermore, the 12-layer and 24-layer variants of our TEK$_F$ approach considerably improve over a comparable \baselineplus~baseline for both Wikipedia (1.9 and 1.1 F1 respectively) and Web (1.6 and 1.0 F1  respectively) indicating that TEK representations are useful even without additional TEK-pretraining. The base variant of our best model \ourmodel, which uses TEK-pretrained TEK-enriched representations records even bigger gains of 3.1 F1 and 2.0 F1 on Wikipedia and Web respectively over a comparable 12-layer \baselineplus~baseline. The 24-layer models show similar trends with improvements of 1.6 and 1.7 F1 over ~\baselineplus. 

% The 12-layer base variant of our \ourmodel~approach, which uses TEK-pretrained TEK-enriched representations, records gains of 3.1 F1 and 2.0 F1 on Wikipedia and Web respectively over a comparable \baselineplus~baseline. We see similar trends for the 24-layer models with improvements of 1.6 F1 and 1.7 F1 over ~\baselineplus. 

\paragraph{MRQA} Table ~\ref{tab:main_results_mrqa} shows in-domain and out-of-domain evaluation on MRQA. 
% An out of domain evaluation of our approach on multiple datasets shows considerable improvements overall with particularly large gains on  BioASQ, DuoRC, and TextbookQA (Table ~\ref{tab:main_results_mrqa}). 
As in the case of TriviaQA, the 12-layer variants of our RoBERTa baselines are competitive with previous work, which includes D-Net~\cite{li2019d} and Delphi~\cite{longpre2019exploration}, the top two systems of the MRQA shared task, while the 24-layer variants considerably outperform the current state of the art across all datasets. \baselineplus~again performs better than \baseline~on all datasets except DROP and RACE. DROP is designed to test arithmetic reasoning, while RACE contains (often fictional and thus not groundable to Wikipedia) passages from English exams for middle and high school students in China. The performance drop after further pretraining on Wikipedia could be a result of multiple factors including the difference in style of required reasoning or content; we leave further investigation of this phenomenon for future work.
% \luanyi{Showing some of the error examples in DROP or RACE would be interesting.} 
% \added{
The base variants of TEK$_F$ and \ourmodel~ outperform both baselines on all other datasets. Comparing the base variant of our full \ourmodel~ approach to \baselineplus, we observe an overall improvement of 1.6 F1 with strong gains on BioASQ (4.2 F1), DuoRC (2.0 F1), and TextbookQA (2.0 F1). The 24-layer variants of \ourmodel~ show similar trends with improvements of 2.1 F1 on BioASQ, 1.1 F1 on DuoRC, and 1.6 F1 on TextbookQA. Our large models see a reduction in the average gain mostly due to drop in performance on DROP. Like in the case of TriviaQA, TEK-pretraining generally improves performance even further where TEK-finetuning is useful (with the exception of DuoRC which sees a small loss of 0.24 F1 due to TEK-pretraining for the large models\footnote{According to the Wilcoxon signed rank test of statistical significance, the large \ourmodel~ is significantly better than TEK$_F$ on BioASQ and TextbookQA $p$-value $<.05$, and is not significantly different from it for DuoRC.}),  with the biggest gains seen on BioASQ.

\paragraph{Takeaways} Both TEK$_{PF}$ and TEK$_F$ record strong gains on benchmarks that focus on factual reasoning outperforming the RoBERTa-based baselines that use only RC-contexts. The success of TEK$_F$ underscores the advantage of \emph{textual} encyclopedic knowledge in that it improves current models even without additional TEK-pretraining. Finally, TEK-pretraining further improves the model's ability to use the retrieved background knowledge for the downstream RC task.
% Finally, the improvements from TEK-enriching task representations are even more prominent for models that have been \ourpretraining~pretrained.

\section{Ablation Studies}
\label{sec:ablation}

\paragraph{TEK vs. Context-only Pretraining}
% \paragraph{Comparing TEK Pretraining and Context-only Pretraining}
\label{sec:pretraining_comparison}
\begin{table}[!t]
  \centering
  \small
  \scalebox{.85}{
  \begin{tabular}{l l l c c c c}
    \toprule
     & Pretraining & Finetuning & {Wiki} & {Web} & {MRQA}\\
    \midrule
    % \multicolumn{4}{c}{Finetuning with background knowledge} \\
    % \midrule
    1 {{RoBERTa++}} & \baselinepluspretraining & \baselinefinetuning & 72.8 & 81.2 & 83.2 \\
    2 {{TEK$_F$}} & \baselinepluspretraining & \ourfinetuning & 74.2 & 82.4 & 83.4 \\
    % \midrule
    3 {{---}} & \ourpretraining & \baselinefinetuning & 72.9 & 81.6 & 83.3 \\
    4  {{~\ourmodel}} & \ourpretraining & \ourfinetuning & \tf{75.1} & \tf{82.8} & \tf{83.7} \\
    \bottomrule
  \end{tabular}
  }
  \caption{Development set F1 on TriviaQA and MRQA for base models using different combinations of pretraining and finetuning. Metrics are average F1 over 3 random finetuning seeds.}
  \label{tab:main_ablations}
\end{table}

% Table \ref{tab:main_ablations} shows that TEK-augmented pretraining further improves the model's ability to use the retreived background knowledge during finetuning (rows 1 and 2). 
% \deleted{To isolate the impact of tailored pretraining with textual knowledge, we compare TEK-pretraining with context-only pretraining (Table \ref{tab:main_ablations}) when representations are enriched with retrieved TEK for the RC task (rows 2 and 4). We see that \ourpretraining~pretraining shows consistent gains across both TriviaQA tasks (+0.9 and +0.4 F1) as well as MRQA (+0.3 F1). This shows that TEK-pretraining further improves the model's ability to use the retrieved background knowledge to improve representations for the downstream RC task.}

% \added{Section~\ref{sec:results} showed how TEK-pretraining further improves the model's ability to use the retrieved background knowledge for the downstream RC task.}
We also compare the two pretraining setups for models which do \emph{not} use background knowledge to form representations for the finetuning tasks. 
Table~\ref{tab:main_ablations} shows results for all four combinations of the pretraining and finetuning method variables, using 12-layer base models on the development sets of TriviaQA and MRQA (in-domain).
Comparing rows 1 and 3, 
we see marginal gains across all datasets for \ourpretraining~pretraining indicating that pretraining with encyclopedic knowledge does not hurt QA performance even when such information is not available during finetuning and inference. While previous work ~\cite{liu2019roberta,joshi2019spanbert} has shown that pretraining with single contiguous chunks of text clearly outperforms BERT's bi-sequence pipeline,\footnote{BERT randomly samples the second sequence from a different document in the corpus with a probability of 0.5.} our results suggest that using \emph{background} sentences from other documents during pretraining has no adverse effect on the downstream tasks we consider.

% Additionally, with rows 3 and 4, we observe that using just 384 tokens of the context window for question and context (\baselinefinetuning-384) consistently performs worse than using all 512 tokens for context (\baselinefinetuning-512). This, taken together with our gains from adding 128 tokens of encyclopedic knowledge to 384 tokens of question and context, suggests that additional context provides benefits complementary to encyclopedic knowledge. 

\begin{table}[!t]
  \centering
  \small
  \begin{tabular}{c c c c c c}
    \toprule
    $N_C$ & $N_B$ & {Wiki} & {Web} &{MRQA}\\
    \midrule 
    384 & 0 & 72.4 & 80.4 & 83.0 \\
    512 & 0 & 72.8 & 81.2 & 83.2 \\
    \midrule
    384 & 128 & \tf{74.2}  & \tf{82.4} & \tf{83.4} \\
    256 & 256 &  73.6 & 82.2 & 83.3\\
    128 & 384 &  68.1 & 79.5 & 81.7\\
    \bottomrule
  \end{tabular}
  \caption{Performance (F1) on TriviaQA and MRQA dev sets for varying lengths of context ($N_C$) and background ($N_B$). All models were finetuned from the same \baselineplus~pretrained checkpoint.}
  \label{tab:knowledge_vs_context}
\end{table}

% Table ~\ref{tab:knowledge_vs_context} examines this tradeoff for varying proportions of context and background for the fixed transformer window size of 512 tokens. We see 

\paragraph{Trade-off between Document Context and Knowledge}
\label{sec:context_vs_background}
Our approach  uses a part of the Transformer window for textual knowledge, instead of additional context from the same document. Having established the usefulness of the background knowledge even without tailored pretraining, we now consider the trade-off between neighboring context and retrieved knowledge (Table ~\ref{tab:knowledge_vs_context}). We first compare using a shorter window of $384$ tokens for RC-contexts 
% (question and passage) 
with using $512$ tokens for RC-contexts (the first two rows). Using longer document context results in consistent gains, some of which our TEK-enriched representations need to sacrifice.
We then consider the trade-off for varying values of context length $N_C$  and background length $N_B$ (rows 2-5). The partitioning of $384$ tokens for context and $128$ for background outperforms other configurations. 
% Moreover, adding up to $256$ tokens of background knowledge improves performance over using only document context. 
\emph{This suggests that relevant encyclopedic knowledge 
% about multiple entities 
from outside of the current document is more useful than long-distance neighboring text from the same document for these benchmarks.}

\paragraph{Choice of the Entity Linker}
\label{sec:linker}
\begin{table}[!t]
  \centering
  \small
  \begin{tabular}{l c c c c}
    \toprule
     & {Wiki} & {Web} &{In} & {Out}\\
     \midrule 
    \baselineplus~ & {71.7} & {81.4} &{83.2} & {68.4} \\
    TEK$_{PF}$ & {76.0} & {83.4} & {83.7} & {70.0} \\
    \midrule
    TEK$_{PF}$-GC & {75.4} & {83.0} & {83.6} & {69.4}\\
    TEK$_{PF}$-TagMe & {75.6} & {83.1} &{83.7} & {69.7} \\
    \bottomrule
  \end{tabular}
  \caption{Performance (F1) of 12-layer \ourmodel~when used with publicly available entity linkers on TriviaQA test sets and MRQA in (In) and out-of-domain (Out).}
  \label{tab:linkers}
\end{table}

Table~\ref{tab:linkers} compares the performance of \ourmodel~when used with publicly available entity linkers, Google Cloud Natural Language API (abbreviated as GC)\footnote{\href{https://cloud.google.com/natural-language/docs/basics\#entity analysis}{https://cloud.google.com/natural-language/docs/basics\#entity analysis}} and TagMe~\cite{ferragina2010tagme}.  Using TagMe results in a minor drop of around 0.3 F1 from \ourmodel~across benchmarks while still maintaining major gains over \baselineplus.  The results indicate that the choice of entity linker can make a difference but our method is robust and performs well with multiple linkers.

% In summary, we find that:
% \begin{enumerate}
% \item Adding textual encyclopedic knowledge  during task-specific finetuning only (i.e., without TEK-pretraining) already significantly improves RC models.
% \item TEK-pretraining further improves the model's ability to use the textual knowledge for end tasks.
% \item TEK-pretraining has no adverse effects even when only document context (without any textual encyclopedic knowledge) is used to represent downstream task inputs.
% \end{enumerate}
\section{Discussion}
\begin{figure}[!ht]
\small
\begin{tabular}{p{7.2cm}}
\toprule
\textbf{Question}: Which river originates in the \emph{Taurus Mountains}, and flows through \emph{Syria and Iraq?} \\
\textbf{Our Answer}: \textcolor{mygreen}{Euphrates}\\
\textbf{Baseline Answer}: \textcolor{red}{Tigris}\\

\textbf{Context}: The Southeastern \emph{Taurus mountains} form the northern boundary... 
% of the Southeastern Anatolia Region and North Mesopotamia. 
They are also the source of the \textcolor{mygreen}{Euphrates} River and \textcolor{red}{Tigris} River. \\
\textbf{Background}: { Originating in eastern Turkey, the \textcolor{mygreen}{Euphrates} flows through \emph{Syria and Iraq}  to join the \textcolor{red}{Tigris}...}
% in the Shatt al - Arab, which empties into the Persian Gulf.
\\
\midrule
\textbf{Question}: What \emph{tyrosine kinase}, involved in a Philadelphia- chromosome positive \emph{chronic myelogenous leukemia}, is the target of \emph{Imatinib (Gleevec)}? \\
\textbf{Our Answer}: \textcolor{mygreen}{BCR-ABL}\\
\textbf{Baseline Answer}: \textcolor{red}{imatinib}\\
\textbf{Context}: \textcolor{red}{Imatinib} induces a durable response in most patients with Philadelphia chromosome-positive \emph{chronic myeloid leukemia}...We show that the only hypothesis consistent with current data on ... gradual decrease in the \textcolor{mygreen}{BCR-ABL} levels seen in most patients is that these patients exhibit a continual, gradual reduction of the LSCs. \\
% This observation may explain the ability to discontinue \textcolor{red}{imatinib} therapy without relapse in some cases. \\
\textbf{Background}: \emph{Chronic myelogenous leukemia} : A 2006 follow up of 553 patients using \textcolor{red}{imatinib} (Gleevec) found an overall survival rate of 89\% after five years. With improved understanding of the nature of the \textcolor{mygreen}{BCR-ABL} protein and its action as a \emph{tyrosine kinase}, targeted therapies (the first of which was \textcolor{red}{imatinib}) that specifically inhibit the activity of the \textcolor{mygreen}{BCR-ABL} protein... 
% protein have been developed.
\\
\midrule
\textbf{Question}: Who did \emph{Germany} defeat to win the \emph{1990 FIFA World Cup}? \\
\textbf{Our Answer}: \textcolor{mygreen}{Argentina} \\
\textbf{Baseline Answer}: \textcolor{red}{Italy} \\
\textbf{Context}: At the \emph{1990 World Cup} in \textcolor{red}{Italy}, West \emph{Germany} won their third World Cup title, 
% in its unprecedented third consecutive final appearance. Captained by Lothar Matthäus, 
defeating Yugoslavia (4-1), UAE 
% (5-1), ...
% the Netherlands (2-1), Czechoslovakia (1-0), and England (1-1, 4-3 on penalty kicks) 
on the way to a final rematch against \textcolor{mygreen}{Argentina}.
% played in the Italian capital of Rome. 
\\
\textbf{Background}: {
At international level, 
He is best known for scoring the winning goal for \emph{Germany} in the \emph{1990 FIFA World Cup} Final against \textcolor{mygreen}{Argentina}...}
% from an 85th - minute penalty kick.}
\\
\midrule
\textbf{Question}: The \emph{state} in which \emph{matter} takes on the \emph{shape} but not the \emph{volume} of its \emph{container} is? \\
\textbf{Our Answer}: \textcolor{mygreen}{Liquid}\\
\textbf{Baseline Answer}: \textcolor{red}{gas}\\
\textbf{Context}:  \textcolor{mygreen}{Liquid} takes the \emph{shape} of its \emph{container}. You could put the same volume of liquid in containers with different shapes. The shape of the liquid in the beaker is short and wide like the beaker, while the shape of the liquid in the graduated cylinder is tall and narrow like that container, but each container holds the same \emph{volume} of liquid... How could you show that \textcolor{red}{gas} spreads out to take the volume as well as the shape of its container? \\
\textbf{Background}: {\textcolor{mygreen}{Liquid} : As such, it is one of the four fundamental \emph{states of matter} is the only state with a definite \emph{volume} but no fixed \emph{shape}.}
% in the Shatt al - Arab, which empties into the Persian Gulf.
\\
\bottomrule
\end{tabular}
    \caption{The first two examples (from TriviaQA and BioASQ) have background knowledge that provides information complementary to the context, while the last two (from TriviaQA and TextbookQA) provides a more direct, yet redundant, phrasing of the information need compared to the original context. }
    \label{fig:analysis_examples}
\end{figure}

% \begin{figure}[!t]
% \centering
% \includegraphics[scale=0.45]{freq.pdf}
% \caption{Performance (F1) of 24-layer TEK$_{PF}$ as a function of answer frequency in Wikipedia on TriviaQA-Wiki.}
% \label{fig:freq_analysis}
% \end{figure}
% Our empirical results show that TEK-enriched representations consistently improve performance across multiple benchmarks. In this section, we analyze the source of these gains.
When are TEK-enriched representations most useful for question answering?
The strongest gains we have seen are on TriviaQA, BioASQ, and TextbookQA. All three datasets involve questions targeting the long tail of factual information, which has sizable coverage in Wikipedia, the encyclopedic collection we use. 
% We found that our largest improvements were on questions where the answer string appeared less than a 1000 times in Wikipedia.
% Figure~\ref{fig:freq_analysis} indicates that most of our gains over~\baselineplus~on TriviaQA Wikipedia are on questions which have relatively infrequent, but not extremely rare, ngrams as answers.
We hypothesize that enriching representations with encyclopedic knowledge could be particularly useful when factual information that might be difficult to ``memorize'' during pretraining is important.
Current pretraining methods are able to store a significant amount of world knowledge into model parameters ~\cite{petroni2019language}; this might enable the model to make correct predictions even from contexts with complex phrasing or partial information. TEK-enriched representations complement this strength via dynamic retrieval of factual knowledge. 
\added{Unlike structured KBs which have been used prominently in previous work, encyclopedic text is more likely to be available for a variety of domains (e.g., biomedical and legal).}
 Improvements on the science-based BioASQ and TextbookQA datasets further suggest that Wikipedia can be used as a \emph{bridge corpus} for more effective domain adaptation for QA.

For 75\% of the examples in the TriviaQA Wikipedia development set where our approach outperforms the context-only baselines, the answer string is mentioned in the background text. A qualitative analysis of these examples indicates that the retrieved background information typically falls into two categories --  (a) where the background helps disambiguate between multiple answer candidates by providing partial pieces of information missing from the original context, and (b) where the background sentences help by providing a redundant but more direct phrasing of the information need compared to the original context. Figure~\ref{fig:analysis_examples} provides examples of each category. 

Even when the retrieved background contains the answer string, our model uses the background only to refine representations of the candidate answers in the original document context; possible answer positions in the background are not considered in our model formulation. This highlights the strength of an encoder with full cross-attention between RC-contexts and background knowledge. The encoder is able to build representations for, and consider possible answers in all document passages, while integrating knowledge from multiple pieces of external textual evidence.

%Does this suggest that the model looks at the background sentences independently of the original context? We see that the model predicts answer spans only from the original context even when it is allowed to predict spans from the background. We posit that it learns to predict spans exclusively from the context since it receives RC supervision from only that part of the input. This, in turn, suggests that the model learns to reason over multiple pieces of information even when it's distributed across the context and the background. This highlights the strength of the cross-encoder approach which creates representations for all of the original context using multiple separate encoder windows, while still being able to encode question-focused background knowledge retrieved dynamically for each context window.

The exact form of background knowledge is dependent on the retrieval function. Our results have shown that contextualizing the input with textual background knowledge, especially after suitable pretraining, improves state of the art methods even with simple entity linking and \emph{ngram}-match retrieval functions.  We hypothesize that more sophisticated retrieval methods could further significantly improve performance (for example, by prioritizing for more complementary information).

%Future work might consider more sophisticated functions which select sentences relevant to the question yet complementary to the given context. 
%\kristout{Could leave something like that for conclusion/future work}\mj{That works too. I really just wanted to frame the contributions so that the retrieval is orthogonal. But feel free to remove/reword.}

% \luanyi{I like this section, it's pretty well written. One minor comment is to address the limitation of relying on a well performed entity linker. Might be better to show an example on RACE where it's not correctly linked (and thus got the wrong answer). This could give people some insight about space of potential improvement. } \mj{I like the idea but maybe we should save this for the camera ready? Just concerned that someone will latch on to this and try to sink the paper.}

\section{Related Work}

% We categorize related work into two main groups: (\textit{i}) integrating background knowledge for language understanding 
% (\textit{ii}) pretraining of general purpose text representations, 
% and (\textit{ii}) question answering.

% \subsection*{Integrating Background Knowledge for Natural Language Understanding (NLU)}
\paragraph{Background Knowledge Integration}
Many NLP tasks require the use of multiple kinds of background knowledge \cite{fillmore, Minsky:1986:SM:22939}. Earlier work~\cite{ratinov2009design,nakashole-mitchell-2015-knowledge} combined features over the given task data with hand-engineered features over knowledge repositories.
% Subsequent work used single-word embeddings such as Glove \cite{pennington-socher-manning:2014:_glove} and Word2vec \cite{mikolov-etal:2013}, trained following the Distributional Hypothesis \cite{Harris1954}. 
% Other forms of external knowledge include using embeddings of background knowledge, predominantly in the form of relation triples from structured knowledge graphs \cite{yang-mitchell-2017-leveraging,bauer-etal-2018-commonsense,mihaylov-frank-2018-knowledgeable,wang-jiang-2019-explicit}.
Other forms of external knowledge include relational knowledge between word or entity pairs, typically integrated via embeddings from structured knowledge graphs (KGs) \cite{yang-mitchell-2017-leveraging,bauer-etal-2018-commonsense,mihaylov-frank-2018-knowledgeable,wang-jiang-2019-explicit} or via word pair embeddings trained from text~\cite{joshi-etal-2019-pair2vec}.
~\citet{weissenborn2017dynamic} used a specialized architecture to integrate background knowledge from ConceptNet and Wikipedia entity descriptions. For open-domain QA, recent works \cite{sun-etal-2019-pullnet,xiong-etal-2019-improving} jointly reasoned over text and KGs, 
% building a graph of text and KG candidate answers, and
via specialized graph-based architectures for defining the flow of information between them. 
% to derive word representations.
These methods did not take advantage of large scale unlabeled text to pre-train deep contextualized representations which have the capacity 
to encode even more knowledge in their parameters. 

Most relevant to ours is work building upon these powerful pretrained representations, and further integrating external knowledge. Recent work focuses on refining pretrained contextualized representations using entity or triple embeddings from structured KGs \cite{peters-etal-2019-knowledge, yang-etal-2019-enhancing-pre, zhang2019ernie}. The KG embeddings are trained separately (often to predict links in the KG), and knowledge from KG is fused with deep Transformer representations via special-purpose architectures. Some of these prior works also pre-train the knowledge fusion layers from unlabeled text through self-supervised objectives~\cite{zhang2019ernie, peters-etal-2019-knowledge}. Instead of separately encoding structured KBs, and then attending  to their single-vector embeddings, we explore directly using wider-coverage textual encyclopedic background knowledge. This enables direct application of a pretrained deep Transformer (RoBERTa) for jointly contextualizing input text and background knowledge. We showed background knowledge integration can be further improved by additional knowledge-augmented self-supervised pretraining. 

\citet{liu2019k} augment text with relevant triples from a structured KB. They process triples as word sequences using BERT with a special-purpose attention masking strategy. This allows the model to partially re-use BERT for encoding and integrating the structured knowledge. Our work uses wider-coverage textual sources instead and shows the power of additional knowledge-tailored self-supervised pretraining. 

% \paragraph{Pretraining Contextualized Representations}
% We have heavily built upon recent work in pretraining general purpose text representations \cite{peters-etal:2018:_deep,radford-etal:2018, devlin-etal-2019-bert, liu2019roberta, joshi2019spanbert} which encode contiguous segments from documents. Our pretrained TEK-enriched representations encode contiguous texts jointly with dynamically retrieved textual encyclopedic knowledge from multiple documents as background. Other pretrained knowledge integration methods \cite{zhang2019ernie, peters-etal-2019-knowledge} refine input text representations by integrating structured KB embeddings instead.
% Combining textual and structured sources is an interesting avenue for future work.

\paragraph{Question Answering}

% We have used a state-of-the-art deep Transformer-based architecture for extractive document-level question answering, also often referred to as a reading comprehension task.

%  For extractive text-based question answering with deep pretrained models, prior work has integrated background knowledge from structured KBs \cite{yang-etal-2019-enhancing-pre,liu2019k}. 
For open-domain QA, where documents known to answer the question are not given as input (e.g. OpenBookQA \cite{mihaylov-etal-2018-suit}), methods exploring retrieval of relevant textual knowledge are a necessity.  Recent work in these areas has focused on improving the evidence retrieval components \cite{lee-etal-2019-latent, banerjee-etal-2019-careful,guu2020realm}, and has used Wikidata triples with textual descriptions of Wikipedia entities as a source of evidence \cite{min2019knowledge}. Other approaches use pseudo-relevance feedback (PRF)~\cite{Xu1996QueryEU} style multi-step retrieval of passages by query reformulation~\cite{buck2017ask,nogueira-cho-2017-task}, entity linking~\cite{das-etal-2019-multi}, and more complex reader-retriever interaction~\cite{das2018multistep}. When multiple candidate contexts are retrieved for open-domain QA, they are sometimes jointly contextualized using a specialized architecture~\cite{min2019knowledge}. We are the first to explore pretraining of representations which can integrate background from multiple documents, and hypothesize that these representations could be further improved by more 
sophisticated retrieval approaches.

\section{Conclusion}
We presented a method to build text representations by jointly contextualizing the input with dynamically retrieved textual encyclopedic knowledge. We showed consistent improvements, in- and out-of-domain, across multiple reading comprehension benchmarks that require factual reasoning and knowledge well represented in the background collection. 

\bibliography{eacl2021}
\bibliographystyle{acl_natbib}

\begin{appendices}
\clearpage
\appendix
\section*{Appendices}
\addcontentsline{toc}{section}{Appendices}
\renewcommand{\thesubsection}{\Alph{subsection}}

\begin{table*}[t]
\begin{center}
\scriptsize
% \begin{tabular}{l|l}
\begin{tabularx}{\textwidth}{X | X }
\toprule
\cls The River Thames known alternatively in parts as the Isis, is a river that flows through southern England including \tf{London}. At 215 miles (346 km), it is the longest river entirely in England and the second-longest in the United Kingdom, after the River Severn. It flows through Oxford (where it is called \tf{the Isis}), Reading, Henley-on-Thames and Windsor. \sep~ \textcolor{blue}{\tf{London} : The city is split by the River Thames into North and South, with an informal central London area in its interior . \sep~\tf{The Isis} : The Isis" is an alternative name for the River Thames, used from its source in the Cotswolds until it is joined by the Thame at Dorchester in Oxfordshire \sep } 
& \cls Which English rowing event is held every year on the River Thames for 5 days ( Wednesday to Sunday ) over the first weekend in July ? \sep~Each year the \tf{World Rowing Championships} is held by FISA ... Major domestic competitions take place in dominant rowing nations and include The Boat Race and \tf{Henley Royal Regatta} in the United Kingdom , the Australian Rowing Championships in Australia , ... \sep~ \textcolor{blue}{ \tf{Henley Royal Regatta} : The regatta lasts for five days ( Wednesday to Sunday ) ending on the first weekend in July . \sep~\tf{World Rowing Championships} : The event then was held every four years until 1974 , when it became an annual competition ... \sep } \\
\bottomrule
\end{tabularx}
\end{center}
\caption{Pretraining (left) and QA finetuning (right) examples which encode contexts with background sentences from Wikipedia. The input is minimally structured by including the source page of each background sentence, and separating the sentences using special \sep~tokens. Background is shown in \textcolor{blue}{blue} and entities are indicated in \tf{bold}.}
\label{tab:pre_vs_fine}
\end{table*}

\subsection{Input Examples}
\label{sec:input_examples}

Table~\ref{tab:pre_vs_fine} shows pretraining (left) and QA finetuning (right) examples which encode contexts with background sentences from Wikipedia.

\subsection{Implementation}
\label{sec:finetune_procedure}
We implemented all models in TensorFlow~\cite{tensorflow2015-whitepaper}. For pretraining, we used the 12-layer RoBERTa-base (125M parameters) and 24-layer RoBERTa-large (355M parameters) configurations, and initialized the parameters from their respective checkpoints. \added{In TEK-augmented pretraining,} we further pretrained the models for 200K steps with a batch size of $512$ and BERT's triangular learning rate schedule with a warmup of $5000$ steps \added{on TEK-augmented contexts}. We used a peak learning rate of $0.0001$ for base and $5e^{-5}$ for large models. 
% For finetuning hyperparameters, see Appendix~\ref{sec:finetune_procedure}. 
All models were trained and evaluated on Google Cloud TPUs. We apply the following finetuning hyperparameters to all methods, including the baselines. For each method, we chose the best model based on dev set performance measured using F1.

\paragraph{TriviaQA\footnote{\href{https://nlp.cs.washington.edu/triviaqa/}{https://nlp.cs.washington.edu/triviaqa/}}}
We follow the input preprocessing of \citet{clark-gardner-2018-simple}. The input to our model is the concatenation of the first four 400-token passages selected by their linear passage ranker.  For training, we define
the gold span to be the first occurrence of the gold
answer(s) in the context~\cite{joshi-etal-2017-triviaqa,talmor-berant-2019-multiqa}.
We choose learning rates from \{1e-5, 2e-5\} and finetune for 5 epochs with a batch size of 32. 

\paragraph{MRQA\footnote{\href{https://github.com/mrqa/MRQA-Shared-Task-2019}{https://github.com/mrqa/MRQA-Shared-Task-2019}}}
We choose learning rates from \{1e-5, 2e-5\} and number of epochs from \{2, 3, 5\} with a batch size of 32. 
For both benchmarks, especially for large models, we found higher learning rates to perform sub-optimally on the development sets. Table ~\ref{tab:hyperparams} reports best performing hyperparamter configurations for each benchmark.

\begin{table}[b]
  \centering
  \small
  \begin{tabular}{l @{\hspace{0.5cm}} c c c}
    \toprule
     Dataset & Epochs & LR \\
    \midrule
    \multicolumn{3}{c}{12-layer Models} \\
    \midrule
    MRQA & 3 & 2e-5 \\
    TriviaQA Wiki & 5 & 1e-5 \\
    TriviaQA Web & 5 & 1e-5 \\
    \midrule
    \multicolumn{3}{c}{24-layer Models} \\
    \midrule
    MRQA & 2 & 1e-5 \\
    TriviaQA Wiki & 5 & 2e-5 \\
    TriviaQA Web & 5 & 1e-5 \\
    \bottomrule
  \end{tabular}
  \caption{Hyperparameter configurations for TEK$_{PF}$}
  \label{tab:hyperparams}
\end{table}
\end{appendices}

\end{document}